\documentclass[11pt,letterpaper]{article}
\usepackage{emnlp2016}
\usepackage{times}
\usepackage{latexsym}

\usepackage{color}
\usepackage[utf8]{inputenc} % allow utf-8 input
\usepackage[T1]{fontenc}    % use 8-bit T1 fonts
\usepackage{hyperref}       % hyperlinks
\usepackage{url}            % simple URL typesetting
\usepackage{booktabs}       % professional-quality tables
\usepackage{amsfonts}       % blackboard math symbols
\usepackage{nicefrac}       % compact symbols for 1/2, etc.
\usepackage{microtype}      % microtypography
\usepackage{graphicx}
\usepackage{epsfig}
\usepackage{natbib}
\usepackage{xspace}

\usepackage{array}
\newcolumntype{L}{>{\arraybackslash}m{10cm}}

\newcommand{\WikiMovies}{{\sc WikiMovies}\xspace}
\definecolor{dgreen}{rgb}{0.0,0.4,0.0}
\definecolor{dred}{rgb}{1,0.0,0.0}
\usepackage{amsmath}
\usepackage{amssymb}

\emnlpfinalcopy

\def\emnlppaperid{679}

\title{Key-Value Memory Networks for Directly Reading Documents}

 \author{
Alexander H. Miller$^{1}$ \mbox{~~}
Adam Fisch$^{1}$ \mbox{~~}
Jesse Dodge$^{1,2}$ \mbox{~~}
Amir-Hossein Karimi$^{1}$ \\
{\bf Antoine Bordes$^{1}$} \mbox{~~}
{\bf Jason Weston$^{1}$} \\
$^{1}$Facebook AI Research, 770 Broadway, New York, NY, USA\\
$^{2}$Language Technologies Institute, Carnegie Mellon University, Pittsburgh, PA, USA\\
{\tt \{ahm,afisch,jessedodge,ahkarimi,abordes,jase\}@fb.com}
 }

\date{}

\begin{document}

\maketitle

\begin{abstract}
Directly reading documents and being able to answer questions from them is an unsolved challenge.
To avoid its inherent difficulty, question answering (QA) has been directed towards
using Knowledge Bases (KBs) instead,
which has proven effective.
Unfortunately KBs often suffer from being too restrictive, as the schema cannot support certain types of answers,
and too sparse, e.g. Wikipedia contains much more information than Freebase.
In this work we introduce a new method, Key-Value Memory Networks,
that makes reading documents more viable
by utilizing different encodings in the addressing  and output stages of the memory read operation.
To compare using  KBs, information extraction or Wikipedia documents directly in a single
framework we construct
 an analysis tool, {\sc WikiMovies}, a QA dataset that contains raw text alongside a preprocessed KB, in the domain of movies.
Our method reduces the gap between all three settings.
It also achieves state-of-the-art results on the existing {\sc WikiQA} benchmark.
\end{abstract}

\section{Introduction}
\vspace{-0.5ex}

%\cite{burger2001issues}
%\cite{fader2014open}
%\cite{voorhees1999trec}

%A huge leap forward in artificial intelligence will be achieved when
%machines will be able to answer any question expressed in natural
%language. As such, q

Question answering (QA) has been a long standing research problem in
natural language processing, with the first systems attempting to
answer questions by directly reading
documents \citep{voorhees2000building}. The development of large-scale Knowledge Bases (KBs) such as Freebase  \citep{bollacker2008freebase}
helped organize information into structured forms, prompting recent progress to focus on answering questions by converting them into logical forms that can be used to query such databases \citep{berant2013semantic,kwiatkowski-EtAl:2013:EMNLP,fader2014open}.

Unfortunately, KBs have intrinsic limitations such as their inevitable incompleteness and fixed schemas that cannot support all varieties of answers.
Since information extraction (IE) \citep{craven2000learning}, intended to
fill in missing information in KBs, is neither accurate nor
reliable enough, collections of raw textual resources and
documents such as Wikipedia will always contain more information.
%than KBs.
%
As a result, even if KBs can be satisfactory for closed-domain problems, they are unlikely
to scale up to answer general questions on any
topic.
Starting from this observation,
%here we propose  to study the problem
in this work we study the problem
of answering by directly reading documents.

Retrieving answers directly from text is harder than
from KBs because information is far less structured, is
indirectly and ambiguously expressed, and is usually scattered across multiple documents.
%
%This explains why, when a satisfactory KB is
%available -- which is typically only the case in closed domains --
%using it instead of raw text is preferred. %, because performance is better.
%
This explains why using a satisfactory KB---typically only available in closed domains---is preferred over raw text.
We postulate that before trying to provide answers that are not in
KBs, document-based QA systems should first reach KB-based systems'
performance in such closed domains, where clear comparison and
evaluation is possible.
To this end, this paper introduces {\sc WikiMovies}, a new
analysis tool that allows for measuring the performance of %loss induced on
QA systems when the knowledge source is switched from a KB to unstructured documents.
{\sc WikiMovies} contains $\sim$100k questions in the movie domain, and was designed
to be answerable by using either a perfect KB
(based on OMDb\footnote{\url{http://www.omdbapi.com}}), Wikipedia pages or an imperfect KB obtained through
running %a standard IE pipeline on those pages.
an engineered IE pipeline on those pages.

To bridge the gap between using a KB and reading documents directly,
we still lack appropriate machine learning algorithms. In this
work we propose the Key-Value Memory Network (KV-MemNN), a new neural network
architecture that generalizes the original Memory Network
\citep{sukhbaatar2015end} and can work with either knowledge source.
The KV-MemNN performs QA by first storing facts in a key-value
structured memory before reasoning on them in order to predict an
answer. The memory is designed so that the model learns to use keys to
address relevant memories with respect to the question, whose corresponding values are subsequently returned.
This structure allows the model to encode prior knowledge for the considered task
and to leverage possibly complex transforms between keys and values,
while still being trained using standard backpropagation via
stochastic gradient descent.

Our experiments on {\sc WikiMovies} indicate that, thanks to its key-value memory,
the KV-MemNN consistently outperforms the
original Memory Network, and reduces the gap between answering from a human-annotated KB,
from an automatically extracted KB or from directly reading Wikipedia.
We confirm our findings on  {\sc WikiQA} \citep{yang2015wikiqa},
another Wikipedia-based QA benchmark where no KB is available,
where we demonstrate that KV-MemNN can reach state-of-the-art results---surpassing
the most recent attention-based neural network models.

%\vspace{-0.25ex}
\section{Related Work}
\vspace{-0.5ex}

Early QA systems were based on information retrieval and were designed
to return snippets of text containing an answer
\citep{voorhees2000building,banko2002askmsr}, with limitations in terms of question
complexity and response coverage.
The creation of large-scale KBs
\citep{auer2007dbpedia,bollacker2008freebase} have led to the
development of a new class of QA methods based on semantic parsing
\citep{berant2013semantic,kwiatkowski-EtAl:2013:EMNLP,fader2014open,yih2015semantic}
that can return precise answers to complicated compositional questions.
Due to the sparsity of KB data, however, the main challenge
shifts from finding answers to developing efficient information
extraction methods to populate KBs automatically
\citep{craven2000learning,carlson2010toward}---not an easy
problem.

For this reason, recent initiatives are returning to the original
setting of directly answering from text using
datasets like {\sc TrecQA} \citep{wang2007jeopardy},
which is based on classical {\sc Trec} resources \citep{voorhees1999trec},
and {\sc WikiQA} \citep{yang2015wikiqa}, which is extracted from Wikipedia.
Both benchmarks are organized around the task of answer sentence
selection, where a system must identify the sentence containing
the correct answer in a collection of documents, but need not return the
actual answer as a KB-based system would do.
Unfortunately, these datasets are very small (hundreds of
examples) and, because of their answer selection setting, do not
offer the option to directly compare answering from a KB against answering from pure text.
Using similar resources as the dialog dataset
of \cite{dodge2015evaluating}, our new benchmark {\sc WikiMovies}
addresses both deficiencies by providing a substantial
corpus of question-answer pairs that can be answered by either using a
KB or a corresponding set of documents.

Even though standard pipeline QA systems like AskMR
\citep{banko2002askmsr} have been recently revisited
\citep{tsai2015web},
the best published results on {\sc TrecQA} and {\sc WikiQA} have been
obtained by either convolutional neural networks
\citep{santos2016attentive,yin2015convolutional,wang2016sentence} or
recurrent neural networks \citep{miao2015neural}---both usually with
attention mechanisms inspired by \citep{bahdanau2014neural}.
In this work, we introduce KV-MemNNs, a Memory Network model that operates a symbolic memory structured as $(key, value)$ pairs.
Such structured memory is not employed
 in any existing attention-based neural network architecture for QA.
As we will show, it gives the model greater
flexibility for encoding knowledge sources
% to retrieve the answers from
and helps shrink the gap between
directly reading documents and answering from a KB.
%versus answering from a KB.

%\citep{hill2015goldilocks}
%\citep{nips15_hermann}

%\citep{bordes2014question}
%\citep{bordes2015large}

%\vspace{-0.5ex}
\section{Key-Value Memory Networks} \label{sec:models}
\vspace{-0.5ex}

\begin{figure*}[t!]
\centerline {
	\epsfxsize=6.5in
	\epsfbox{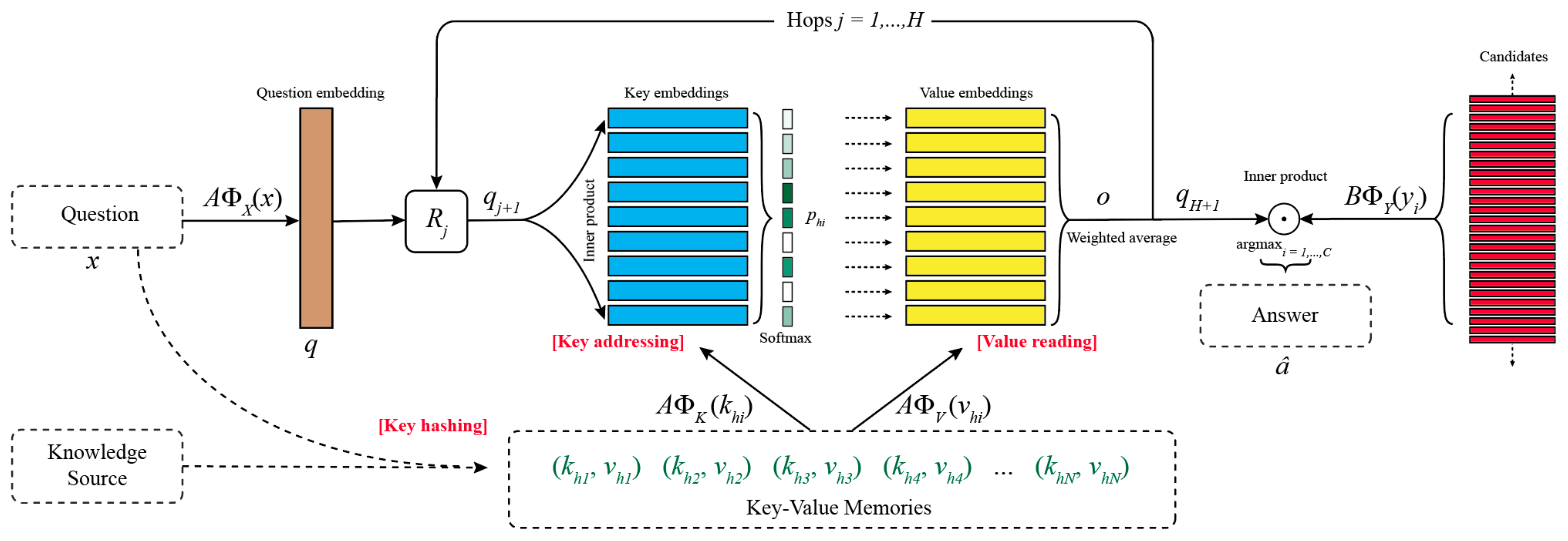}
  }
  \caption{\label{fig:KV-Memnn_diagram} The Key-Value Memory Network model for question answering. See Section~\ref{sec:models} for details.}
\end{figure*}

%We demonstrate its usefulness when applied to the problem of reading documents and
%answering questions about them.
% which we apply to the problem of reading documents.
The Key-Value Memory Network model is based on the  Memory Network (MemNNs)
model \citep{Weston14,sukhbaatar2015end} which has proven useful for a variety of
document reading and question answering tasks:
for reading children's books and answering questions about them \citep{hill2015goldilocks},
for complex reasoning over simulated stories \citep{weston2015towards}
and for utilizing KBs to answer questions \citep{bordes2015large}.

Key-value paired memories are a generalization of the way context (e.g. knowledge bases or
documents to be read) are stored in memory. The lookup (addressing) stage is based on the key
memory while the reading stage (giving the returned result) uses the value memory.
%In contrast conventional MemNNs use an
This gives both (i) greater flexibility for the practitioner to encode prior knowledge
about their task; and (ii) more effective power in the model via nontrivial
transforms between key and value.
The key should be designed with features to help match it to the question,
while the value should be designed with features to help match it to the response (answer).
An important property of the model is that the entire model
can be trained with key-value transforms
while still using standard backpropagation via stochastic gradient descent.

\if 0
Memory Networks \citep{Weston14,sukhbaatar2015end}
 are a recently introduced model that have proved useful for a variety of
document reading and question answering tasks:
for reading children's books and answering questions about them \citep{hill2015goldilocks},
for complex reasoning over simulated stories \citep{weston2015towards},
and for utilizing KBs to answer questions \citep{bordes2015large}.
\fi

\subsection{Model Description}

Our model is based on
 the end-to-end Memory Network  architecture of \cite{sukhbaatar2015end}.
%which
%is a flexible class of models for storing and retrieving the relevant context needed to answer
%questions that can be trained end-to-end via stochastic gradient descent.
A high-level view of both models is as follows:
one defines a memory, which is a possibly very large array of slots
 which can encode both long-term
and short-term context.
%At test time one is given a query (e.g. the question in QA tasks), which is used
%to iteratively address and read from the memory (these iterations are
%also referred to as ``hops'')
 %looking for relevant information to answer the question.
% where the iterations are termed ``hops''.
%Finally, after collecting information from the memory,
%the original query and the retrieved context are combined as features to predict the final response.
At test time one is given a query (e.g. the question in QA tasks), which is used to iteratively address and read from the memory (these iterations are also referred to as ``hops'') looking for relevant information to answer the question. At each step, the collected information from the memory is cumulatively added to the original query to build context for the next round. At the last iteration, the final retrieved context and the most recent query are combined as features to predict a response from a list of candidates.

%We will first describe this architecture and then detail the simple, but very useful
%extensions to this method developed in this work.
%
Figure~\ref{fig:KV-Memnn_diagram} illustrates the KV-MemNN model architecture.

In KV-MemNNs we define  the memory slots as pairs of vectors $(k_1, v_1)\dots, (k_M, v_M)$
and denote the question $x$.
%Following \cite{dodge2015evaluating}
The addressing and reading of the memory
 involves three steps:
\begin{itemize}
\item {\bf Key Hashing:}
 the question can be
 used to pre-select a small subset of the possibly large array. This is done
using an inverted index that finds a subset $(k_{h_1},v_{h_1}), \dots, (k_{h_N},v_{h_N})$
 of memories of size $N$ where the key shares at
least one word with the question with frequency < $F=1000$ (to ignore stop words),
 following \cite{dodge2015evaluating}.
%Clearly
More sophisticated
retrieval schemes could be used here, see e.g. \cite{manning2008introduction},
%but this simple scheme sufficed for our needs in the
%following experiments.
\item {\bf Key Addressing:} during addressing, each candidate memory is assigned a relevance probability by comparing the question to each key:
\[
  p_{h_i} =  \text{Softmax}(A\Phi_X(x) \cdot A\Phi_K(k_{h_i}))
\]
where $\Phi_{\cdot}$ are feature maps of dimension $D$, $A$ is a $d \times D$ matrix
and  $\text{Softmax}(z_i) = e^{z_i} / \sum_j e^{z_j}$.
We discuss choices of feature map in Sec. \ref{sec:featuremap}.

\item {\bf Value Reading:} in the final reading step, the values of the memories are read by taking their weighted sum using the addressing probabilities, and the vector $o$ is returned:
\[
       o = \sum_i p_{h_i}  A\Phi_V(v_{h_i})~.
\]
\end{itemize}

The memory access process is conducted  by the ``controller'' neural network
 using $q = A \Phi_X(x)$ as the query.
After receiving the result $o$, the query is updated with $q_2 = R_1(q + o)$ where $R$ is a $d \times d$ matrix. The memory access is then repeated (specifically, only the addressing and reading steps, but not the hashing), using a different matrix $R_{j}$ on each hop, $j$. The key addressing equation is transformed accordingly to use the updated query:
\[
  p_{h_i} =  \text{Softmax}(q_{j+1}^\top A\Phi_K(k_{h_i}))~.
\]

The motivation for this is that new evidence can be combined into the query to focus on and
retrieve more pertinent information in subsequent accesses.
Finally, after a fixed number $H$ hops, the resulting state of the controller is used to compute a final prediction over the possible outputs:
\[
    \hat{a} = \text{argmax}_{i={1,\dots,C}} \text{Softmax}(q_{H+1}^\top B \Phi_Y(y_i))
\]
where $y_i$ are the possible candidate outputs, e.g. all the entities in the KB,
or all possible candidate answer sentences in the case of a dataset like {\sc WikiQA}
(see Sec. \ref{sec:wikiqa}).
The $d \times D$ matrix $B$ can also be constrained to be identical to $A$.
 The whole network is trained end-to-end, and the model
learns to perform the iterative accesses to output the desired target $a$
 by minimizing a standard cross-entropy
loss between $\hat{a}$ and the correct answer $a$.
Backpropagation and stochastic gradient descent are thus used to learn the matrices
$A, B$ and $R_1, \dots, R_H$.

To obtain the standard End-To-End Memory Network of \cite{sukhbaatar2015end}
one can simply set the key and value to be the same for all memories.
Hashing was not used in that paper,
%  (as well as
%remove the hashing stage from memory addressing). Key hashing is important for computational
but is important for computational
efficiency for large memory sizes, as already shown
 in \cite{dodge2015evaluating}.
We will now go on to describe specific applications of
 key-value memories for the task of reading KBs or documents.

\subsection{Key-Value Memories} \label{sec:featuremap}

There are a variety of ways to employ key-value memories %for reading documents
that can have important effects on overall performance.
The ability to encode prior knowledge in this way is an important
 component of KV-MemNNs, and we are free to define $\Phi_X, \Phi_Y, \Phi_K$ and $\Phi_V$
for the query, answer, keys and values respectively.
We now describe several possible variants of $\Phi_K$ and $\Phi_V$
that we tried in our experiments,
for simplicity we kept $\Phi_X$ and  $\Phi_Y$ fixed as bag-of-words representations.

\paragraph{KB Triple}
Knowledge base entries have a structure of triple ``subject {\em relation} object'' (see Table~\ref{fig:blade} for examples).
The representation we consider is simple:
the key is composed of the left-hand side entity (subject) and the relation,
and the value is the right-hand side entity (object).
We double the KB and consider the reversed relation as well
(e.g. we now have two triples ``Blade Runner {\em directed\_by} Ridley Scott'' and
``Ridley Scott {\em !directed\_by} Blade Runner'' where {\em !directed\_by} is a
different entry in the dictionary than {\em directed\_by}). Having the entry both
ways round is important for answering different kinds of questions
(``Who directed Blade Runner?'' vs. ``What did Ridley Scott direct?'').
For a standard MemNN that does not have key-value pairs the whole triple has to
be encoded into the same memory slot.

\paragraph{Sentence Level}
For representing a document, one can
%
%Documents are
 split it up into sentences, with each memory slot encoding one
sentence.  Both the key and the value encode the entire sentence as a
bag-of-words.  As the key and value are the same in this case, this
is identical to a standard MemNN and this approach has been used in
several papers \citep{weston2015towards,dodge2015evaluating}.

\paragraph{Window Level}
Documents are split up into windows of $W$ words; in our tasks we only include
windows where the center word is an entity. Windows are represented using bag-of-words.
Window representations for MemNNs have been shown to work well previously
\citep{hill2015goldilocks}.
However, in Key-Value MemNNs we encode the key as the entire window, and the value
as only the center word, which is not possible in the MemNN architecture.
This makes sense because the entire window is more likely to be pertinent as a match
for the question (as the key), whereas the entity at the center is more pertinent
as a match for the answer (as the value). We will compare these approaches in our
experiments.

\paragraph{Window + Center Encoding}
Instead of representing the window as a pure bag-of-words, thus mixing the window center
with the rest of the window, we can also encode them with different features. Here,
we double the size, $D$, of the dictionary and encode the center of the window and the value
 using the second dictionary. This should help the model pick out the relevance
of the window center (more related to the answer)
as compared to the words either side of it (more related to the question).

\paragraph{Window + Title}
The title of a document is commonly the answer to a question that relates to the text
it contains. For example ``What did Harrison Ford star in?'' can be (partially) answered by the
Wikipedia document with the title ``Blade Runner''. For this reason, we also consider
a representation where the key is the word window as before,
but the value is the document title.
We also keep all the standard (window, center) key-value pairs from the window-level
representation as well, thus doubling the number of memory slots in comparison.
To differentiate the two keys with different values we add an extra feature
``\_window\_'' or ``\_title\_'' to the key, depending on the value.
The ``\_title\_'' version also includes the actual movie title in the key.
This representation can be combined with center encoding. Note that this representation is inherently specific to datasets in which there is an apparent or meaningful title for each document.

%\vspace{-1ex}
\section{The WikiMovies Benchmark} \label{sec:data}
\vspace{-0.5ex}
The \WikiMovies ~benchmark consists of question-answer pairs in the domain of movies.
It was built with the following goals in mind:
(i) machine learning techniques should have ample training examples for learning;
and (ii) one can analyze easily the performance of different representations of knowledge and break
down the results by question type.
The dataset can be downloaded from \url{http://fb.ai/babi}.

\newcolumntype{L}{>{\arraybackslash}m{10cm}}

\definecolor{dblue}{rgb}{0.0,0.0,0.6}
\definecolor{dred}{rgb}{0.3,0.0,0.0}
\definecolor{die}{rgb}{0.6,0.6,0.0}
\definecolor{dgreen}{rgb}{0.0,0.6,0.0}

%\newcolumntype{L}{>{\arraybackslash}m{10cm}}

\begin{table}[t]
\begin{small}
\begin{center}
 \resizebox{1\linewidth}{!}{
 {
%\begin{tabular}{l}
\begin{tabular}{|L|}
\hline
{\bf Doc: Wikipedia Article for Blade Runner (partially shown)}\\
\vspace{1mm}
\textcolor{dblue}{Blade Runner is a 1982 American neo-noir dystopian science fiction film
 directed by Ridley Scott and starring Harrison Ford, Rutger Hauer, Sean Young, and Edward James Olmos. The screenplay, written by Hampton Fancher and David Peoples, is a modified film adaptation of the 1968 novel ``Do Androids Dream of Electric Sheep?''  by Philip K. Dick. The film depicts a dystopian Los Angeles in November 2019 in which genetically engineered replicants, which are visually indistinguishable from adult humans, are manufactured by the powerful Tyrell Corporation as well as by other ``mega-corporations'' around the world.   %\dots}\\
Their use on Earth is banned and replicants are exclusively used for dangerous, menial, or leisure work on off-world colonies. Replicants who defy the ban and return to Earth are hunted down and ``retired'' by special police operatives known as ``Blade Runners''. \dots}
\\
\hline
%\end{tabular}
%\begin{tabular}{|L|}
{\text{\bf KB entries for Blade Runner (subset)}}\\
\vspace{1mm}
\textcolor{red}{Blade Runner {\em directed\_by} Ridley Scott}\\
\textcolor{red}{Blade Runner {\em written\_by} Philip K. Dick, Hampton Fancher}\\
\textcolor{red}{Blade Runner {\em starred\_actors} Harrison Ford, Sean Young, \dots} \\% Edward James Olmos\\
\textcolor{red}{Blade Runner {\em release\_year} 1982}\\
\textcolor{red}{Blade Runner {\em has\_tags} dystopian, noir, police, androids, \dots}
%dystopia, cult film, police, future, \dots %harrison ford, library, national film registry, philip k. dick, los angeles, ridley scott, androids, noir, visual, 2, rutger hauer, dystopian, edward james olmos, director's cut, sean young, android\\
\\
%\textcolor{red}{\dots} \\
\hline
{\text{\bf IE entries for Blade Runner (subset)}}\\
\vspace{1mm}
\textcolor{die}{Blade Runner, Ridley Scott {\em directed}    dystopian, science fiction, film}\\
\textcolor{die}{Hampton Fancher {\em written}    Blade Runner}\\
%\textcolor{die}{Blade Runner {\em brought}    Philip K. Dick}\\
\textcolor{die}{Blade Runner {\em starred}   Harrison Ford, Rutger Hauer, Sean Young\dots}\\ % Edward James Olmos}\\
\textcolor{die}{Blade Runner {\em labelled}    1982 neo noir}\\
\textcolor{die}{special police, Blade {\em retired} Blade Runner}\\
\textcolor{die}{Blade Runner, special police {\em known} Blade}
\\
\hline
{\bf Questions for Blade Runner (subset)}\\
\vspace{1mm}
\textcolor{dgreen}{Ridley Scott directed which films?}\\
\textcolor{dgreen}{What year was the movie Blade Runner released?}\\
\textcolor{dgreen}{Who is the writer of the film Blade Runner?}\\
   %What films can be described by ridley scott?\\
   %Which films can be described by sean young?\\
   %Which movies are about edward james olmos?\\
   %What movies are about harrison ford?\\
   %Which movies are about noir?\\
   %What films can be described by android?\\
\textcolor{dgreen}{Which films can be described by dystopian?}\\
   %Which movie did Hampton Fancher write?\\
\textcolor{dgreen}{Which movies was Philip K. Dick the writer of?}\\
\textcolor{dgreen}{Can you describe movie Blade Runner in a few words?}
\\
\hline
\end{tabular}
}}
\caption{
\label{fig:blade}
%{\bf WikiMovies}: Questions and KB, Wikipedia sources.}
{\bf \WikiMovies}: Questions, Doc, KB and IE sources.}
\end{center}
\end{small}
\vspace{-1ex}
\end{table}

%The dataset has around 100,000 question-answer pairs that are split between train, dev and test sets.
%This is much larger than most existing datasets, for example the WikiQA dataset \citep{yang2015wikiqa}
% for which we also conduct experiments in Sec. \ref{sec:wikiqa} has only $\sim$1000 training pairs.
%Being able to separate the problem of severe overfitting from the type of knowledge representation..

\subsection{Knowledge Representations} \label{sec:kr}

We construct three forms of knowledge representation:
(i) Doc: raw Wikipedia documents consisting of the pages of the movies mentioned;
(ii) KB: a classical graph-based KB consisting of entities
and relations created from the Open Movie Database (OMDb) and MovieLens;
and (iii) IE: information extraction performed on the Wikipedia pages to
build a KB in a similar form as (ii).
We take care to construct QA pairs such that they are all potentially answerable
from either the KB from (ii) or the original  Wikipedia documents from (i) to
eliminate data sparsity issues. However, it should
 be noted that the advantage of working from raw documents in real applications
is that data sparsity is less of a concern than for a KB, while on the other hand the KB
has the information already parsed in a form amenable to manipulation by machines.
This dataset can help analyze what  methods we need
to close the gap between all three settings, and in particular what
are the best methods for reading documents when a KB is not available.
A sample of the dataset is shown in Table~\ref{fig:blade}.

\paragraph{Doc}
We selected a set of Wikipedia articles about movies
by identifying a set of movies from OMDb\footnote{\tiny{\url{http://beforethecode.com/projects/omdb/download.aspx}}}
that had an associated article by title match.
%We identified a set of movies from OMDb\footnote{Downloaded from \tiny{\url{http://beforethecode.com/projects/omdb/download.aspx}}.}
%that had an associated Wikipedia article by title match,
We keep the title and the first section (before the contents box) for each article.
This gives $\sim$17k documents (movies) which comprise the set of documents our
models will read from in order to answer questions.

\paragraph{KB}
Our set of movies were also matched to
the MovieLens dataset\footnote{\tiny{\url{http://grouplens.org/datasets/movielens/}}}.
We built a KB using OMDb and MovieLens metadata with entries for each movie and nine different relation types:
director, writer, actor, release year, language, genre, tags, IMDb rating and IMDb votes,
with $\sim$10k related actors, $\sim$6k directors and
$\sim$43k  entities in total.
The KB is stored as triples; see Table~\ref{fig:blade} for examples.
% such as
 %{\small{ {\sc (Young Frankenstein, starred\_actors, Gene Wilder)}}} and
%{\small {\sc (The Little Mermaid, has\_tags, Disney Animation)}}.
IMDb ratings and votes
are originally real-valued but are binned and converted to text
 (``unheard of'', ``unknown'', ``well known'', ``highly watched'', ``famous'').
We finally only retain KB triples where the entities also appear in the Wikipedia
articles\footnote{The dataset also
includes the slightly larger version without this constraint.}
to try to
guarantee that all QA pairs will be equally answerable by either the KB or Wikipedia document
sources.

\paragraph{IE} As an alternative to directly reading documents,
we explore leveraging information extraction  techniques to
transform documents into a KB format.
%Constraining the memories solely to facts identified by an IE system introduces a few tradeoffs.
%Processing each Wikipedia entry into a series of semi-structured facts mimics
%some of the attractive attributes of the KB,
An IE-KB representation has attractive properties
such as more precise and compact expressions of facts
and logical key-value pairings based on subject-verb-object groupings.
This can come at the cost of lower recall due to malformed or completely missing triplets.
For IE we use standard open-source software followed by some task-specific
engineering to improve the results.
We first employ coreference resolution via the Stanford NLP Toolkit \citep{manning2014stanford} to reduce ambiguity by replacing pronominal (``he'', ``it'') and nominal (``the film'') references with their representative entities. Next we use the SENNA semantic role labeling tool \citep{senna_collobert} to uncover the grammatical structure of each sentence and pair verbs with their arguments. Each triplet is cleaned of words that are not recognized entities,
and lemmatization is done to collapse different inflections of important task-specific verbs to one form (e.g. stars, starring, star $\rightarrow$ starred).
Finally, we append the movie title to each triple similar to the ``Window + Title''
representation of Sec. \ref{sec:featuremap}, which improved results.

\subsection{Question-Answer Pairs}
%\paragraph{QA Pairs}
%The dataset has more than 100,000 question-answer pairs.
%Being able to separate the problem of severe overfitting from the type of knowledge representation..
%
Within the dataset's more than 100,000 question-answer pairs, we distinguish 13
classes of question corresponding to different kinds of edges in our KB.
They range in scope from specific---such as
{\em actor to movie}:~``What movies did Harrison Ford star in?'' and
{\em movie to actors}:~``Who starred in Blade Runner?''---to more general,
such as {\em tag to movie}:~``Which films can be described by {\em dystopian}?'';
see Table \ref{table:breakdown} for the full list.
For some question there can be multiple correct answers.
%
%The topics correspond to different edges in our KB, and range in scope from specific ({\em movie to actors}  -- ``Who starred in Blade Runner?'') to more general ({\em tag to movie} -- ``Which films can be described by {\em dystopian}?''). For each question type there is a set of possible answers.
%
%as shown in Table \ref{table:breakdown}.
%he topics correspond to different edges in our KB, and range in scope from specific ({\em movie to actors}  -- ``Who starred in Blade Runner?'') to more general ({\em tag to movie} -- ``Which films can be described by {\em dystopian}?'').
%For each question type there is a set of possible answers.
%corresponding to different kinds of edges in our KB:
%{\em actor to movie} (``What movies did Harrison Ford star in?''),
%{\em movie to actors} (``Who starred in Blade Runner?''),
%{\em movie to director}, {\em director to movie},
%{\em movie to writer}, {\em writer to movie},
%{\em movie to tags}, {\em tag to movie},
%{\em movie to year}, {\em movie to genre}, {\em movie to language},
%{\em movie to IMDb rating} and
%{\em movie to IMDb votes}.

Using SimpleQuestions \citep{bordes2015large},
an existing open-domain question answering dataset based on Freebase,
we identified the subset of questions posed by human annotators that covered
our question types.
%We expanded this set to cover all of our KB by substituting the entities
%in those questions to also apply them to other questions.
We created our question set by substituting the entities in those questions
with entities from all of our KB triples.
For example, if the original question written by an annotator was
``What movies did Harrison Ford star in?'', we created a pattern
``What movies did [@actor] star in?'', which we substitute for any
other actors in our set, and repeat this for all annotations.
%We removed {\em tag to movie} questions with more than 50 answers,
%and
We split the questions into disjoint training, development and test sets
with $\sim$96k, 10k and 10k examples, respectively.
The same question (even worded differently) cannot appear in
both train and test sets.
Note that this is much larger than most existing datasets;
for example, the {\sc WikiQA} dataset \citep{yang2015wikiqa}
for which we also conduct experiments
in Sec. \ref{sec:wikiqa} has only $\sim$1000 training pairs.

%\vspace{-0.5ex}
\section{Experiments}\label{sec:exp}
\vspace{-0.5ex}
\begin{table}[t!]
	\begin{center}
	\label{tab:WikiMoviesres}
   	\resizebox{1\linewidth}{!}{{
	\renewcommand{\arraystretch}{1.0}
    	\begin{tabular}{l|c|c|c|}
      		Method &  KB   &  IE  & Doc \\
      		\hline
		\citep{bordes2014question} QA system & 93.5 & 56.5 & N/A \\
		Supervised Embeddings         & 54.4 & 54.4 & 54.4 \\
		Memory Network                    & 78.5 & 63.4 & 69.9 \\ %69.9
		Key-Value Memory Network  & \textbf{93.9} & \textbf{68.3} & \textbf{76.2} \\ % 73.9
    	\end{tabular}
    	}}
    	\caption{
          \label{table:main-res}
          { Test results (\% hits@1)
on \WikiMovies, comparing human-annotated KB (KB), information extraction-based KB (IE),
and directly reading Wikipedia documents (Doc).}}
  	\end{center}
%  \vspace{-3ex}
\end{table}

\begin{table}[t!]
	\begin{center}
    	\resizebox{1\linewidth}{!}{{
	\renewcommand{\arraystretch}{1.0}
    	\begin{tabular}{l|c|c|c|}
      		Memory Representation  &   Doc \\
      		\hline
		Sentence-level               &  52.4 \\  % dev-kv:52.4        dev-memnn:?
		Window-level                 &  66.8 \\  % dev-kv:66.77 test-kv:66.4   dev-memnn:?
		Window-level + Title      &  74.1 \\  % dev-kv:74.1 test-kv:73.9      dev-memnn:?
		Window-level + Center Encoding + Title & \textbf{76.9} \\ % dev-kv:76.9 test-kv:76.2 dev-memnn:?
    	\end{tabular}
    	}}
    	\caption{
          \label{table:memkv-res}
{Development set performance (\% hits@1) with different document memory representations for KV-MemNNs. }}
  	\end{center}
%  \vspace{-3ex}
\end{table}

This section describes our experiments %with KV-MemNNs
 on \WikiMovies and
{\sc WikiQA}.

\subsection{WikiMovies} \label{sec:WikiMovies}

We conducted experiments on the \WikiMovies~ dataset described
in Sec. \ref{sec:data}. Our main goal is to
compare the performance of KB, IE and Wikipedia (Doc) sources when
trying varying learning methods.
%and the ability of different learning methods on them.
We compare four approaches:
(i) the QA system of
\cite{bordes2014question} that performs well on existing datasets
WebQuestions \citep{berant2013semantic} and SimpleQuestions \citep{bordes2015large} that use KBs only; % \citep{bordes2015large},
(ii) supervised embeddings that do not make use of a KB at all
but learn question-to-answer embeddings directly
and hence act as a sanity check \citep{dodge2015evaluating};
(iii) Memory Networks; and (iv) Key-Value
Memory Networks.
Performance is reported using the accuracy of the top hit (single answer)
over all possible answers (all entities), i.e. the hits@1 metric measured in percent.
In all cases hyperparameters are optimized on the development set, including
the memory representations of Sec. \ref{sec:featuremap} for MemNNs and KV-MemNNs.
As MemNNs do not support key-value pairs, we concatenate key and value together
when they differ instead.
%
%The best choice of hops was $H=1$ for MemNNs and $H=2$ for KV-MemNNs.

The main results are given in Table \ref{table:main-res}.
The  QA system of \cite{bordes2014question} outperforms Supervised Embeddings
and Memory Networks for KB and IE-based KB representations, but is designed
to work with a KB, not with documents (hence the N/A in that column).
%Key-Value Memory Networks outperform Memory Networks and all other methods
%on all three data source types.
However, Key-Value Memory Networks outperform all other methods
on all three data source types.
Reading from Wikipedia documents directly (Doc) outperforms an IE-based KB (IE),
which is an encouraging result towards automated machine reading though a
gap to a human-annotated KB still remains (93.9 vs. 76.2).
The best memory representation for directly reading
documents uses ``Window-level + Center Encoding + Title''
($W=7$ and $H=2$);
see Table \ref{table:memkv-res} for a comparison of results for different
representation types.
Both center encoding and title features help the window-level representation, while
sentence-level is inferior.

\begin{table}[t!]
	\begin{center}
    	\resizebox{0.75\linewidth}{!}{{
    	\begin{tabular}{l|c|c|c|}
      		Question Type  & KB & IE & Doc \\
      		\hline
		Writer to Movie           &  97  &  72  &  91  \\
		Tag to Movie               &  85  &  35  &  49  \\
		Movie to Year             &  95   &  75  &  89  \\
		Movie to Writer           &  95   &  61  &  64  \\
		Movie to Tags             &  94   &  47  &  48  \\
		Movie to Language     &  96   &  62  &  84  \\
		Movie to IMDb Votes   &  92   &  92  &  92  \\
		Movie to IMDb Rating  &  94   &  75  &  92  \\
		Movie to Genre           &  97   &  84  &  86  \\
		Movie to Director        &  93   &  76  &  79  \\
		Movie to Actors           &  91   &  64  &  64  \\
		Director to Movie        &  90   &  78  &   91  \\
		Actor to Movie            &  93   &   66  &  83  \\
    	\end{tabular}
    	}}
    	\caption{
	\label{table:breakdown}
{Breakdown of test results (\% hits@1) on \WikiMovies for
Key-Value Memory Networks
%KV-MemNNs
using different
knowledge representations.}}
 	\end{center}
%  \vspace{-3ex}
\end{table}

\paragraph{QA Breakdown}
A breakdown by question type comparing the different data sources for KV-MemNNs is given
in Table \ref{table:breakdown}. IE loses out especially to Doc (and KB) on
Writer, Director and Actor to Movie, perhaps because coreference is difficult in these cases --
although it has other losses elsewhere too. Note that only 56\% of
subject-object pairs in IE match the triples in the original KB, so losses are expected.
Doc loses out to KB particularly on Tag to Movie, Movie to Tags, Movie to Writer and
Movie to Actors. Tag questions
are hard because they can reference more or less any word in the entire
Wikipedia document; see Table \ref{fig:blade}. Movie to Writer/Actor are hard
because there is likely only one or a few references to the answer across all documents, whereas
for Writer/Actor to Movie there are more possible answers to find.

\begin{table}[t!]
	\begin{center}
          \begin{small}
  	\begin{tabular}{l|c|c|}
      	    Knowledge Representation   &  KV-MemNN  \\
     	    \hline
       	KB                               &   93.9    \\
      	One Template Sentence            &   82.9    \\
       	All Templates Sentences          &   80.0        \\
       	One Template + Coreference      &   76.0         \\
       	One Template + Conjunctions       &   74.0        \\
       	All Templates + Conj. + Coref.   &   72.5       \\
       	Wikipedia Documents              &    76.2   \\
  	 \end{tabular}
 %   	}}
    	\caption{
	\label{tab:templateres}
              { Analysis of test set results (\% hits@1) for KB vs. Synthetic Docs on \WikiMovies.}}
          \end{small}
  	\end{center}
%  \vspace{-3ex}
\end{table}

\paragraph{KB vs. Synthetic Document Analysis}
%\textcolor{red}{TODO: talk about templates (or not)}
To further understand the difference between using a KB versus reading documents directly,
we conducted an experiment where we constructed synthetic documents using the KB.
For a given movie, we use a simple grammar to construct a synthetic ``Wikipedia'' document
based on the KB triples: for each relation type we have a set of template phrases
(100 in total)  used to generate the fact, e.g.
``Blade Runner came out in 1982'' for the entry {\sc Blade Runner release\_year 1982}.
We can then parameterize the complexity of our synthetic documents:
(i) using one template, or all of them;
(ii) using conjunctions to combine facts into single sentences or not;
and (iii) using coreference between sentences where we replace the movie name with ``it''.\footnote{This data is also part of the \WikiMovies benchmark.}
The purpose of this experiment is to find which aspects are responsible for the gap
in performance to a KB.
The results are given in Table \ref{tab:templateres}.
They indicate that some of the loss (93.9\% for KB to 82.9\% for One Template Sentence)
 in performance is due directly to representing in sentence form, making the subject, relation
and object harder to extract.
Moving to a larger number of templates does not deteriorate performance much (80\%).
 The remaining performance drop seems to be split roughly
equally between conjunctions (74\%) and coreference (76\%).
%When combined, which is the hardest synthetic dataset (All Templates + Conj. + Coref.), this
The hardest synthetic dataset combines these (All Templates + Conj. + Coref.) and
is actually harder than using the real Wikipedia documents (72.5\% vs. 76.2\%).
This is possibly because the amount of conjunctions and coreferences we make are artificially
too high (50\% and 80\% of the time, respectively).

\begin{table}[t!]
	\begin{center}
    	\resizebox{1\linewidth}{!}{{
	\begin{tabular}{l|c|c|}
      		Method & MAP & MRR \\
      		\hline
		Word Cnt                                                 &  0.4891 & 0.4924 \\
		Wgt Word Cnt                                          &  0.5099 & 0.5132 \\
		2-gram CNN    \citep{yang2015wikiqa}    & 0.6520 & 0.6652 \\
		AP-CNN  \citep{santos2016attentive}      & 0.6886 & 0.6957  \\
		Attentive LSTM \citep{miao2015neural}          & 0.6886  & 0.7069 \\
		Attentive CNN \citep{yin2015convolutional}    & 0.6921  & 0.7108 \\
		L.D.C. \citep{wang2016sentence}  & 0.7058  & 0.7226 \\
		\hline
      		Memory Network                     &       0.5170  &       0.5236  \\
      		Key-Value Memory Network   & {\bf 0.7069} & {\bf 0.7265} \\
    	\end{tabular}
    	}}
    	\caption{
	\label{tab:wikiqares}
              { Test results on WikiQA.}}
  	\end{center}
%  \vspace{-3ex}
\end{table}

\subsection{WikiQA} \label{sec:wikiqa}

{\sc WikiQA} \citep{yang2015wikiqa} is an existing dataset for answer sentence selection
using Wikipedia as the knowledge source. The task is, given a question, to select
the sentence coming from a Wikipedia document that best answers the question,
where performance is measured using mean average precision (MAP) and mean reciprocal rank (MRR)
of the ranked set of answers. The dataset uses a pre-built
information retrieval step and hence provides a fixed set of candidate sentences per question,
 so systems do not have to consider ranking all of Wikipedia.
In contrast to \WikiMovies, the training set size is small ($\sim$1000 examples) while
the topic is much more broad (all of Wikipedia, rather than just movies) and the
questions can only be answered by reading the documents, so no comparison to the use
of KBs can be performed. However, a wide range of methods have already been tried on
{\sc WikiQA}, thus providing a useful benchmark to test if the same results
found on \WikiMovies carry across to {\sc WikiQA}, in particular the performance
of Key-Value Memory Networks.

Due to the size of the training set, following many
other works \citep{yang2015wikiqa,santos2016attentive,miao2015neural}
we pre-trained the word vectors (matrices $A$ and $B$ which are constrained to be identical)
before training KV-MemNNs.
We employed Supervised Embeddings \citep{dodge2015evaluating}
for that goal, training on all of Wikipedia while
treating the input as a random sentence and the target as the subsequent sentence.
We then trained KV-MemNNs with dropout regularization:
we sample words from the question, memory representations and the answers,
choosing the dropout rate using the development set.
Finally, again following other successful methods \citep{yin2015convolutional},
we combine our approach
with exact matching word features between question and answers.
Key hashing was not used as candidates were already pre-selected.
To represent the memories, we used the Window-Level representation (the best choice on
the dev set was $W=7$) as the key and the whole sentence as the value, as the value should match the answer which in this case is a sentence.
Additionally, in the representation
all numbers in the text and the phrase ``how many'' in the question
were replaced with the feature ``\_number\_''.
The best choice of hops was also $H=2$ for KV-MemNNs.

The results are given in Table \ref{tab:wikiqares}.
Key-Value Memory Networks outperform a large set of other methods,
although the results of the L.D.C. method of \citep{wang2016sentence} are very similar.
Memory Networks, which cannot easily pair windows to sentences, perform much worse,
highlighting the importance of key-value memories.

%\vspace{-0.5ex}
\section{Conclusion}
\vspace{-0.5ex}

%Conclusion!

%key,value for graph, duh duh duh
%]

We studied the problem of directly reading documents in order to answer questions,
concentrating our analysis on the gap between such direct methods and using
human-annotated or automatically constructed KBs.
We presented a new model, Key-Value Memory Networks, which helps bridge this gap,
outperforming several other methods across two datasets, \WikiMovies and {\sc WikiQA}.
However, some gap in performance still remains. \WikiMovies serves as an
 analysis tool to shed some light on the causes.
Future work should try to close this gap further.

Key-Value Memory Networks are versatile models for reading documents or KBs and answering
questions about them---allowing to encode prior knowledge about the task at hand
in the key and value memories. These models could be applied to storing and
reading memories
for other tasks as well, and future work should try them in other domains,
such as in a full dialog setting.
%dialog,
%for example dialgog

\bibliography{dialog}
\bibliographystyle{natbib}

\end{document}